\documentclass{article}
\usepackage{spconf,amsmath,graphicx}
\usepackage{times}
\usepackage{epsfig}
\usepackage{graphicx}
\usepackage{amsmath}
\usepackage{amsfonts}
\usepackage{amssymb}
\usepackage{color}
\usepackage{times}
\usepackage{epsfig}
\usepackage{amssymb}
\usepackage{graphics} % for pdf, bitmapped graphics files
\usepackage{mathptmx} % assumes new font selection scheme installed
\usepackage{mathrsfs}
\usepackage{multirow}
\usepackage{longtable}
\usepackage{rotating}
\usepackage{array}
\usepackage{float}
\usepackage{subfigure}
\usepackage{algorithm}
\usepackage{algorithmic}
\usepackage{caption}
\usepackage{xcolor}
\usepackage{mathrsfs}
\usepackage[colorlinks,linkcolor=black,citecolor=black]{hyperref}
\usepackage{indentfirst}
\usepackage{booktabs}
\usepackage{setspace}
\usepackage[english]{babel}
\usepackage{amsmath} % used for boldsymbol.

\title{CMF: Cascaded Multi-model Fusion for Referring Image Segmentation}
\name{Jianhua Yang\(^{1}\), Yan Huang\(^{2}\), Zhanyu Ma\(^{1,3}\), Liang Wang\(^{2,4,5}\)\thanks{This work was jointly supported by National Key Research and Development Program of China Grant No. 2018AAA0100400, National Natural Science Foundation of China (61525306, 61633021, 61721004, 61806194, U1803261, 61976132, 61922015, 61773071, and U19B2036), Beijing Nova Program (Z201100006820079), Shandong Provincial Key Research and Development Program (2019JZZY010119), Key Research Program of Frontier Sciences CAS Grant No. ZDBS-LY-JSC032, CAS-AIR, and in part by Beijing Natural Science Foundation Project under Grant Z200002.}}
\address{\(^{1}\) PRIS Lab.,
	   School of Artificial Intelligence, Beijing University of Posts and Telecommunications\\
	\(^{2}\)Center for Research on Intelligent Perception and Computing (CRIPC), National Laboratory of\\
	 Pattern Recognition (NLPR), Institute of Automation, Chinese Academy of Sciences (CASIA)\\
    \(^{3}\)Beijing Academy of Artificial Intelligence, Beijing, China\\
    \(^{4}\)Artificial Intelligence Research, Chinese Academy of Sciences (AIR-CAS)\\
    \(^{5}\)School of Artificial Intelligence, University of Chinese Academy of Sciences}

\begin{document}

\captionsetup{font={small}}

%\ninept
%
\maketitle
\begin{abstract}
In this work, we address the task of referring image segmentation (RIS), which aims at predicting a segmentation mask for the object described by a natural language expression. Most existing methods focus on establishing unidirectional or directional relationships between visual and linguistic features to associate two modalities together, while the multi-scale context is ignored or insufficiently modeled. Multi-scale context is crucial to localize and segment those objects that have large scale variations during the multi-modal fusion process. To solve this problem, we propose a simple yet effective Cascaded Multi-modal Fusion (CMF) module, which stacks multiple atrous convolutional layers in parallel and further introduces a cascaded branch to fuse visual and linguistic features. The cascaded branch can progressively integrate multi-scale contextual information and facilitate the alignment of two modalities during the multi-modal fusion process. Experimental results on four benchmark datasets demonstrate that our method outperforms most state-of-the-art methods. Code is available at https://github.com/jianhua2022/CMF-Refseg.

%We have proposed an effective cascaded multi-modal fusion (CMF) module for referring image segmentation. It stacks multiple atrous convolutional layers in parallel and further introduces a cascaded branch to fused visual and linguistic features using these layers with gradually increased dilated rates. The model can iteratively integrate multi-scale context and facilitate the alignment of two modalities during multi-modal fusion process. We perform experiments on four benchmark datasets and achieve state-of-the-art performance.

% It requires the model to handle the alignment of different semantic concepts between language and vision.
%Multi-scale visual context is crucial to recognize and localize these objects that have large scale variations for referring image segmentation. Previous works ignore the importance of multi-scale context in the fusion process and directly use ASPP as a decoder to predict the mask from multi-modal features. The simple yet effective

\end{abstract}
\vspace{-0.3cm}
\section{Introduction}
\label{sec:intro}
\vspace{-0.3cm}

%(1) introduce the importance of context modeling

\begin{figure}
	\centering	
	%\vspace{-0.2}
	\includegraphics[width=7.0cm]{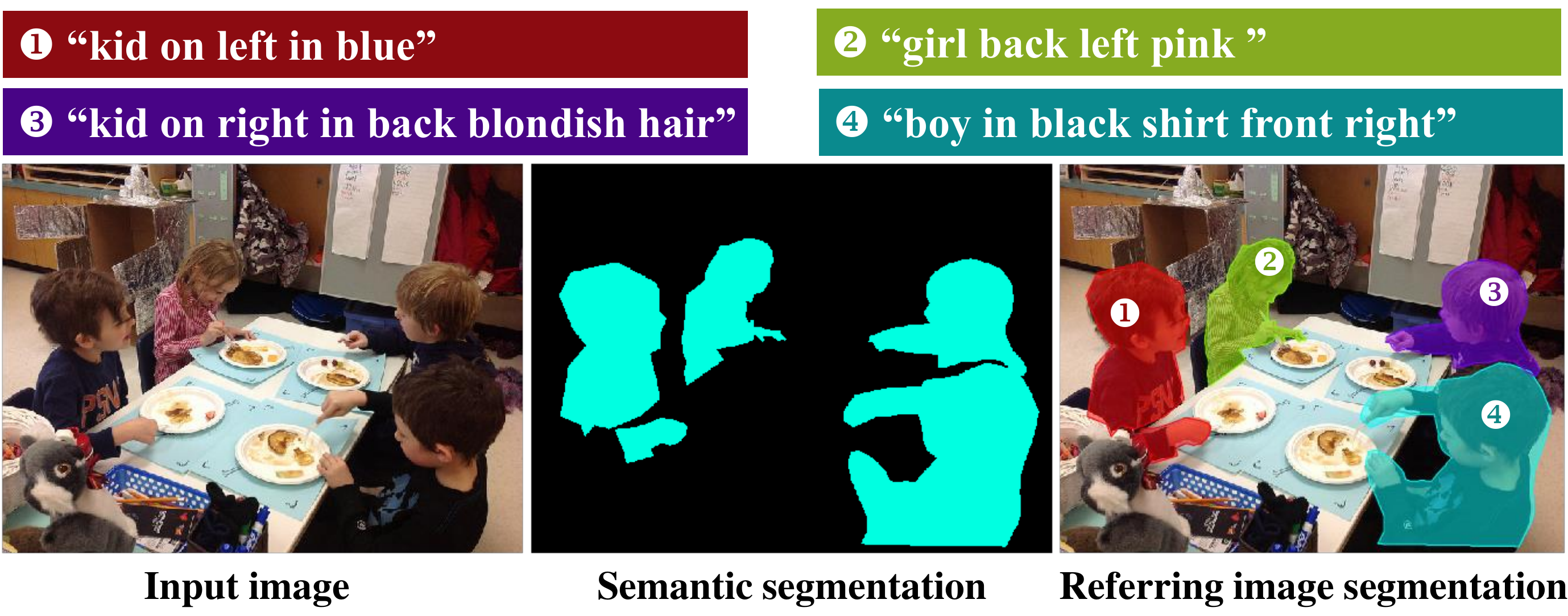}
	\vspace{-0.3cm}
	\caption{Illustration of referring image segmentation, which aims at segmenting the object specified by a natural language expression.}
	\label{fig:illustration}
	\vspace{-0.7cm}
\end{figure}

Referring image segmentation (RIS) \cite{hu2016segmentation} is a challenging task which associates semantics of natural language expressions with contents of images. As illustrated in Fig.\ref{fig:illustration}, given an image and a natural language expression, the goal of RIS is to predict a binary mask for the object specified by the expression. The categories of semantic segmentation are predefined and fixed, while RIS can take arbitrary expressions as input to describe the object of interest. The expressions may contain diverse semantic concepts, such as entities, attributes and actions. Thus, how to align these semantic concepts with visual contents is a main challenge of this task.

%RIS has attracted much attention recently as its applications on human-computer interactions and interactive image editing.

%\vspace{-0.4cm}

%(2) previous work on context modeling and existing problem

Previous works \cite{hu2016segmentation,liu2017recurrent,li2018referring} usually concatenated visual and linguistic features extracted from the convolutional neural network (CNN) and long short-term memory (LSTM) \cite{hochreiter1997long} network respectively, and then predicted the mask from integrated multi-modal features via a fully convolutional network (FCN) \cite{long2015fully}. These works ignored the linguistic variations of input expressions, and treated each word equally to each visual region. Thus it is hard to accurately distinguish the referred object from the background. Some works applied dynamic filters \cite{margffoy2018dynamic} or word attention \cite{ye2020dual} to learn adaptive expression representations but lacked interactions between vision and language. Interactions between two modalities can associate each word with each image region, and further highlight the features of the target object for accurate segmentation. Recent works proposed to model the interactions by introducing unidirectional \cite{shi2018key,chen2019see} or bi-directional \cite{hu2020bi} attention mechanisms.

%Recent works applied dynamic filters \cite{margffoy2018dynamic} or word attention \cite{ye2020dual} to learn adaptive expression representation but lack interactions between vision and language. To model the interactions between two modalities, KWAN \cite{shi2018key} and STEP \cite{chen2019see} applied vision guided linguistic attention to associate each image region with all words of an expression, and obtained adaptively linguistic context for each vision region. BRINet \cite{hu2020bi} further introduced language guided visual attention to learn bi-directional interactions between vision and language. Aforementioned works model the inter- or intra-modal interactions individually, CMSA \cite{ye2019cross} used self-attention on multimodal features to capture the intra- and inter-modal interactions simultaneously.

Although promising results have been achieved in these works, the multi-scale visual context in the multi-modal fusion process has not been explored. Multi-scale context modeling has verified its effectiveness on boosting the segmentation accuracy of objects in semantic segmentation \cite{chen2017deeplab,liu2015parsenet,zhao2017pyramid,fu2019dual}. Recent works also have shown that the performance of RIS can be further improved through aggregating long-range context from concatenated visual and linguistic features \cite{ye2019cross} with self-attention \cite{vaswani2017attention}, or collecting multi-scale context from fused multi-model features \cite{hu2020bi,luo2020cascade} with atrous spatial pyramid pooling (ASPP) \cite{chen2017deeplab,chen2017rethinking}. However, the former is high memory cost for computing the affinity map and may introduce redundant features, which are harmful to distinguish the referred object. The latter captures multi-scale context after two modalities are fused. The fused multi-modal features may contain heterogeneous noises from two modalities, which result in the model cannot adaptively learn the scale variations of objects. Different from these works, we argue that the multi-scale contextual information is important to facilitate the alignment of two modalities and correctly localize the referred object during the multi-modal fusion process.

%Although promising results has been achieved by establishing interactions between two modalities in these works, the multi-scale visual context in multi-modal fusion process has not been exploited well. Visual context modeling has verified its effectiveness on boosting the segmentation accuracy of objects in semantic segmentation \cite{chen2017deeplab,liu2015parsenet,zhao2017pyramid,fu2019dual}. Recent approaches of RIS have shown that through aggregating contextual information from multi-modal features, the segmentation performance can be further improved. Specifically, CMSA \cite{ye2019cross} applied self-attention to aggregate features from all positions of the feature map by learning an affinity matrix. While the dense computation of affinity map may lead to high memory cost and introduce redundant features which are harmful to distinguish the referred objects.

%CGAN \cite{luo2020cascade} and BRINet \cite{hu2020bi} focused on building complex interactions to highlight the features of referred objects on multi-modal features, then fed the highlighted features into an atrous spatial pyramid pooling (ASPP) \cite{chen2017rethinking,chen2017deeplab} to model multi-scale context and predict the segmentation masks.

%Compared with self-attention, ASPP is more cost-effective and indeed improves the segmentation performance by stacking multiple parallel atrous convolutional layers with different dilated rates. However, multi-scale context modeling is not further explored and discussed in these works.

In this paper, we focus on multi-scale context modeling during vision and language fusion process rather than after fused multi-modal features. Specifically, we propose a Cascaded Multi-modal Fusion (CMF) module, which can effectively aggregate multi-scale contextual information into multi-modal fusion process and encourage two modalities alignment at each position of fused feature map. The module is based on ASPP and further introduces a cascaded multi-modal fusion branch, where we iteratively fuse visual and linguistic features with atrous convolutional layers with gradually increased dilated rates. Besides, to integrate the fused multi-modal features from different layers of visual backbone for segmentation mask refining, we introduce a bi-directionally convolutional gated recurrent unit (GRU) \cite{chung2014empirical} to fuse them from top-down and bottom-up paths. Finally, we conduct extensive experiments on four benchmark datasets to validate our proposed method. Experimental results show that our method outperforms most state-of-the-art methods.

\vspace{-0.3cm}
\section{Related Work}
\label{sec:relatedwork}
\vspace{-0.3cm}

% introduction, then stat-of-the-art methods ...

\textbf{Referring Image Segmentation (RIS)}. RIS aims at segmenting the object specified by a natural language expression. Hu \emph{et al.} \cite{hu2016segmentation} made the first effort for this task via a direct concatenation of visual and linguistic features from CNN and LSTM. In order to associate visual regions with individual words and highlight features of the target object, unidirectional and bi-directional interactions between visual and linguistic features have been explored by attention mechanism \cite{shi2018key,ye2020dual,ye2019cross,hu2020bi,chen2019see,luo2020cascade}. Multi-level feature aggregation \cite{li2018referring,ye2019cross,hu2020bi}, generative adversarial learning \cite{qiu2019referring}, and query reconstruction \cite{2020Query} were also investigated to refine multi-modal features and further improve the performance of segmentation. However, these works ignore or ineffectively model multi-scale context during multi-modal fusion process, where the context is crucial to align two modalities and accurately segment the objects with large scale variations. Different from them, our work focuses on multi-scale context modeling in an effective way during the multi-modal fusion process.

\textbf{Context Modeling}. Modeling the multi-scale contextual information plays a key role in semantic segmentation. PSPNet \cite{zhao2017pyramid} collected context from different scales via a spatial pyramid pooling (SPP) module. ASPP \cite{chen2017deeplab,chen2017rethinking} stacked multiple parallel atrous convolutional layers with different dilated rates to capture multi-scale context. DANet \cite{fu2019dual} utilized a position attention and a channel attention with self-attention mechanism to aggregate long-range context. These approaches are proposed to solve the semantic segmentation rather than RIS task. Our work aims at modeling multi-scale context to improve the segmentation accuracy of referred objects in RIS task, which is more challenging than semantic segmentation.

%The context modeling is also crucial for RIS task, CMSA \cite{ye2019cross} proposed to aggregate global contextual information with self-attention but the memory cost is expensive. Other works \cite{luo2020cascade,hu2020bi} directly used ASPP to capture multi-scale context from fused multi-modal features. While these works pay more attention to obtain more discriminated multi-modal features and ignore the importance of multi-scale context modeling.

%Different from them, our work focus on context modeling using ASPP and further proposed a cascaded ASPP fusion module.
% DenseASPP followed the idea of dense connection to encode contextual information in a dense way.
% low level features containing detailed information to refine local parts of segmentation results.
%ParseNet \cite{liu2015parsenet} adopted global pooling feature as global context and concatenated with original feature map.

% figure 1
\begin{figure}[!t]
	\centering
	\includegraphics[width=8.0cm]{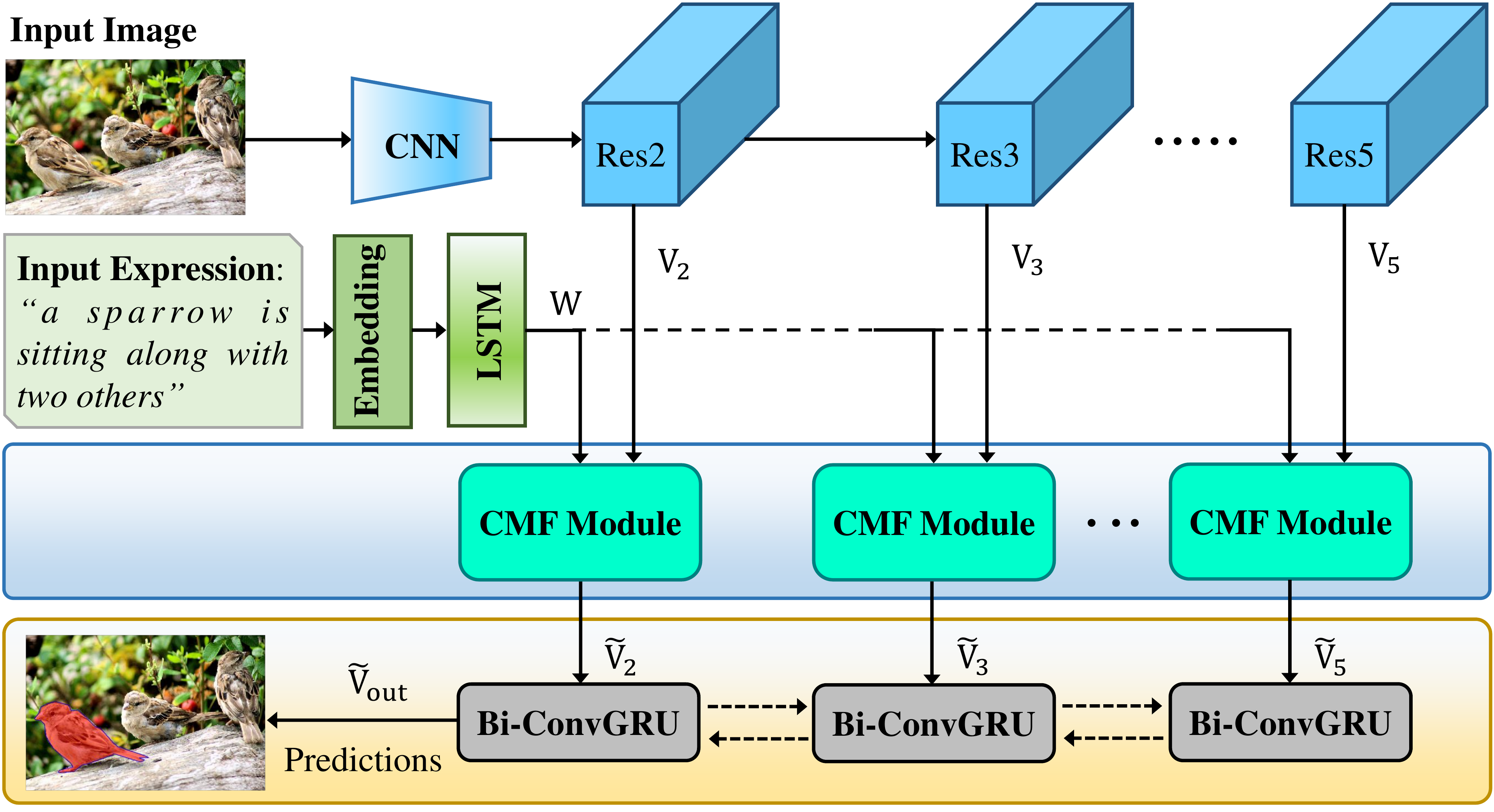}
	\vspace{-0.2cm}
	\caption{Illustration of the overall architecture. It consists of three parts: vision and language encoders, Cascaded Multi-modal Fusion (CMF) module and a Bi-ConvGRU.}
	\vspace{-0.6cm}
	\label{fig:framework}
\end{figure}

%Our method consists of three parts: (1) The backbone CNN and LSTM are applied to extract visual and linguistic features, respectively. (2) Cascaded Multi-modal Fusion (CMF) module is used to iteratively integrate contextual information and fuse two modalities. (3) Bi-ConvGRU is introduced to refine multi-modal features from different levels and predict segmentation masks.

\vspace{-0.3cm}
\section{The Proposed Method}
\label{sec:framework}
\vspace{-0.3cm}

The overall architecture of our proposed model is illustrated in Fig. \ref{fig:framework}. The visual and linguistic features are firstly extracted using a CNN and a one-layer LSTM respectively, and then fed into Cascaded Multi-modal Fusion (CMF) module to fuse them. Finally, a Bi-ConvGRU is introduced to refine the fused multi-modal features from different layers of CNN. The details will be elaborated in the following.

%The overall architecture of our proposed model is illustrated in Fig.\ref{fig:framework}. Specifically, we first extract visual and linguistic features utilizing a CNN and an one-layer LSTM, respectively. Then the visual and linguistic features are fed into the proposed Cascaded Multi-modal Fusion (CMF) module to progressively fuse and conduct multi-scale context modeling. Finally, a Bi-directional convolutional GRU (Bi-ConvGRU) is introduces to refine multi-modal features from different scales and predict the segmentation mask.

\vspace{-0.3cm}
\subsection{Visual and Linguistic Feature Extraction}
\vspace{-0.3cm}
Our model takes an image and a referring expression as inputs. The visual features $V_{l}$ (\emph{l}=2,3,4,5) are extracted from the corresponding layers of \emph{Res2}, \emph{Res3}, \emph{Res4} and \emph{Res5} in the CNN backbone. Then we transform these features into the same size, \emph{i.e.}, $V_{l} \in{\mathbb{R}^{h\times{w}\times{D_{v}}}}$. For an expression with \emph{T} words, we embed each word with the Glove \cite{pennington2014glove} , and then employ a one-layer LSTM to encode the word embedding sequence. Finally, all words in an expression are represented with their corresponding LSTM hidden states and denoted as $L \in{\mathbb{R}^{T\times{D_{w}}}}$. Following prior works \cite{hu2016segmentation,liu2017recurrent}, we also introduce an 8-D spatial coordinate feature $S \in{\mathbb{R}^{h\times{w}\times{8}}}$ to enhance the position reasoning of the model.

\vspace{-0.3cm}
\subsection{Cascaded Multi-modal Fusion (CMF) Module}
\vspace{-0.3cm}

As prior works \cite{hu2016segmentation,liu2017recurrent,li2018referring}, the heterogeneous features from two modalities are concatenated and followed by an $1\times{1}$ convolutional layer to fuse them. However, such a simple multi-modal fusion manner ignored the linguistic variations of input expressions and treated each pixel individually without considering the multi-scale contextual information during the multi-modal fusion process. Thus, we propose a CMF module to fuse two modalities by taking linguistic variations and multi-scale contextual information into account.

% figure 2
\begin{figure}[!t]
	\centering
	\includegraphics[width=8.0cm]{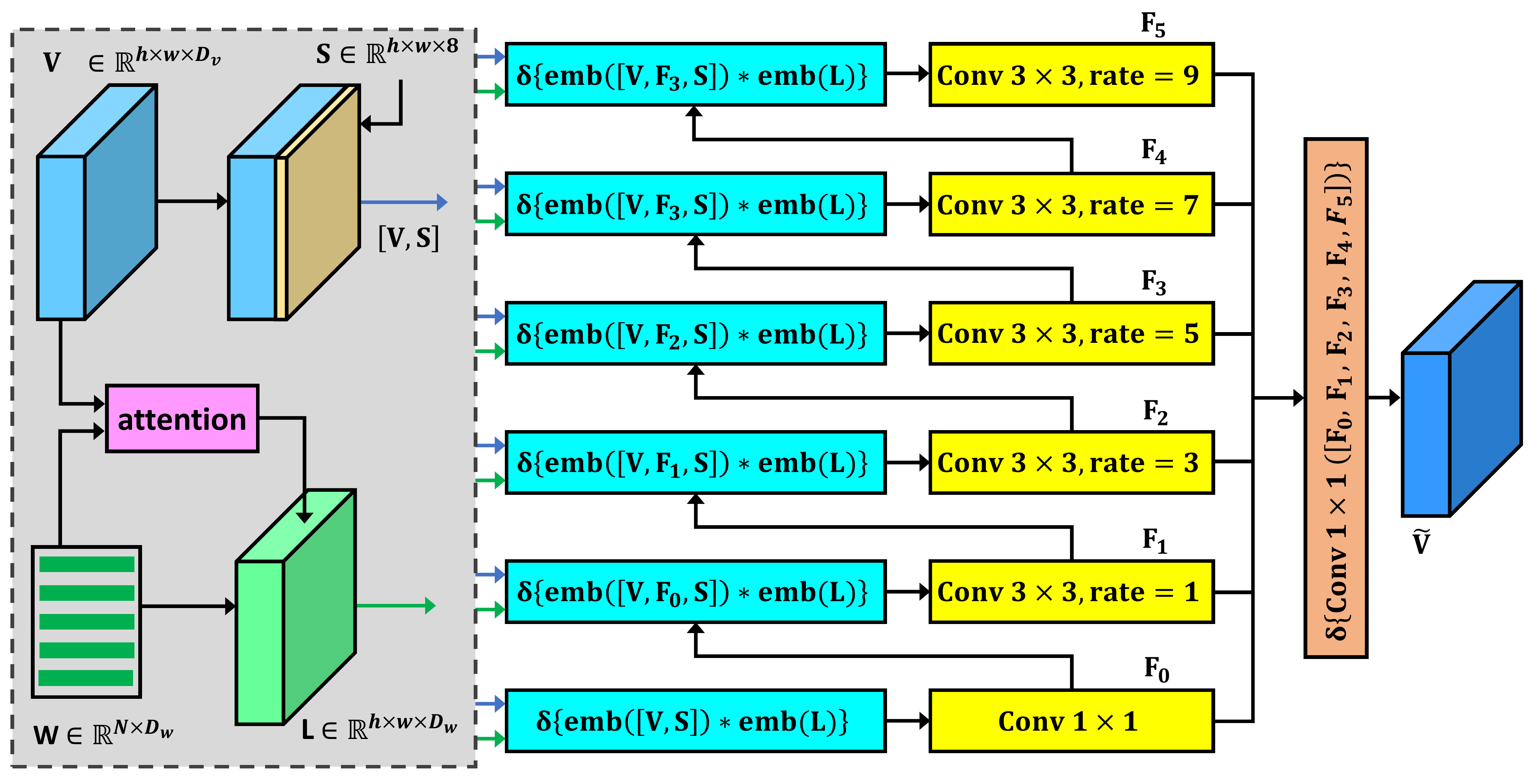}
	\vspace{-0.2cm}
	\caption{The cascaded multi-modal fusion (CMF) module. $emb(\cdot)$ denotes $1\times{1}$ convolutional layer, $\delta(\cdot)$ is ReLU function, $\ast$ means Hadamard product.}
	\vspace{-0.6cm}
	\label{fig:cascaded_aspp}
\end{figure}

The details of the CMF module is shown in Fig.\ref{fig:cascaded_aspp}. Concretely, to learn more robust linguistic representation and eliminate the effect of linguistic variations, we first introduce an image-to-word attention to compute the relevance between each word and each visual region, then utilize the calculated attention matrix followed by a softmax layer to compute the weighted summation of originally linguistic features. For the \emph{i}-th visual region and the \emph{t}-th word, the attention matrix is calculated by:
\vspace{-0.3cm}
\begin{equation}
\vspace{-0.2cm}
A_{i,t} = softmax[(W_{v}v_{i})^{T}(W_{h}h_{t})],
\vspace{-0.1cm}
\end{equation}
then the updated linguistic representation $L$ for the visual feature map $V$ is $L = \{l_{i}\} \in{\mathbb{R}^{h\times{w}\times{D_{w}}}}$, where $l_{i}$ is calculated by:
\vspace{-0.2cm}
\begin{equation}
\vspace{-0.2cm}
l_{i} = \sum\nolimits_{t=1}^{T}A_{i,t}h_{t}.
\end{equation}

To capture multi-scale contextual information, the original ASPP \cite{chen2017deeplab,chen2017rethinking} stacks multiple parallel $3\times{3}$ convolutional layers with different dilated rates and further fuses them with concatenation followed by an $1\times{1}$ convolutional layer. To effectively model multi-scale context and facilitate alignment of two modalities for RIS task, we introduce a cascaded fusion branch on original ASPP, as shown in Fig.\ref{fig:cascaded_aspp}. This branch first fuses two modalities with an $1\times{1}$ convolutional layer, and then iteratively fuses two modalities using atrous convolutional layers with gradually increased dilated rates. More specifically, for the \emph{n}-th fusion layer in the cascaded branch, we first concatenate the input visual feature $V$, spatial feature $S$ and the fusion result $F_{n-1}$ from the $(n-1)$-th fusion layer, which can be denoted as $[V, S, F_{n-1}]$. Then we apply two linearly embedding layers (\emph{i.e.}, $1\times{1}$ convolutional layers) to transform the concatenated features and updated linguistic representation into the same dimension. Instead of concatenation similar to previous works \cite{hu2016segmentation,liu2017recurrent,li2018referring}, here we combine two types of features with Hadamard product followed by a non-linear projection. Finally, we feed the combined multi-modal features into a $3\times{3}$ convolutional layers with a specific dilated rate to fuse them. By introducing such a cascaded fusion module, the visual contextual information is gradually integrated into multi-modal fusion process in cascaded and parallel directions of CMF module.

\vspace{-0.3cm}
\subsection{Multi-level Feature Fusion and Mask Prediction}
\vspace{-0.3cm}

Previous approaches \cite{li2018referring,ye2019cross,chen2019see,hu2020bi} have shown that integrating multi-modal features from different levels of CNN can further improve the accuracy of segmentation masks. In our work, we introduce a bi-directionally convolutional GRU (Bi-ConvGRU) to progressively integrate the fused multi-modal features in bottom-up and top-down manners, which is corresponding to the two directions of forward and backward paths. The top-down manner can enhance the lower-level features with rich semantics, while the bottom-up manner can compensate spatial details for higher-level features with lower-level ones. The output of the Bi-ConvGRU is formulated as:
\vspace{-0.4cm}
\begin{equation}
\vspace{-0.3cm}
\widetilde{V}_{out} = ReLU( W_{p}^{\overrightarrow{H}} \overrightarrow{H_{l}} + W_{p}^{\overleftarrow{H}} \overleftarrow{H_{l}} + b),
\end{equation}
where $\overrightarrow{H}$ and $\overleftarrow{H}$ denote the last hidden states of two directions, $b$ is the bias term, and $\widetilde{V}_{out} \in{\mathbb{R}^{h\times{w}\times{D_{o}}}}$ is the integrated multi-modal features. Finally, the same decoder layers and binary cross-entropy loss as previous works \cite{li2018referring,hu2020bi} are adopted to predict the segmentation mask and optimize the network, respectively.

%This can be attribute to that higher-level visual features contain more semantic information which benefits to recognize and localize the target object, while lower-level visual features preserve more local and detailed information which is useful to refine segmentation masks. Different from them,

%Therefore, in our work, multiple CMF modules are firstly used to fuse linguistic features and different level visual features. Then we introduce a bi-directional convolutional GRU (Bi-ConvGRU) to iteratively fuse the multi-modal features in bottom-up and top-down manners, which is corresponding to the two directions of forward and backward paths.

% The feature representation F obtained from Equ. is specific to a particular layer in CNN. previous work has shown that fusing features at multiple scales can improve the performance of referring image segmentation.
% integrate multi-level features from to further refine the segmentation mask. Inspired by them, we introduce...
%we use both bottom-up and top-down manner to guide the multi-level feature fusion gradually.
% the bottom-up pathway, higher-level features provide the global and semantic guidance to the lower-level ones.
% the top-down pathway, we hope that the lower-level features provide the local and fine guidance to the higher-level ones.
% in this network, we employ BconvGRU to encode the multi-modal features from different levels. Bi-ConvGRU process the input data into two directions of forward and backward paths.
% the output of the BconvGRU is calculated as

\vspace{-0.3cm}
\section{Experiments}
\label{sec:experiments}
\vspace{-0.3cm}

\subsection{Experiment Settings}
\vspace{-0.3cm}

\textbf{Datasets and Protocols}. We perform experiments on four benchmark datasets, including ReferIt \cite{kazemzadeh2014referitgame}, G-Ref \cite{mao2016generation}, UNC \cite{yu2016modeling} and UNC+ \cite{yu2016modeling}. The ReferIt contains 130,525 expressions for 96,654 regions in 19,894 images, the categories of regions are objects or stuff (\emph{e.g.}, ``sky", ``wall"). The G-Ref consists of 26,711 images with 104,560 expressions for 54,822 objects. The average length of expressions (8.4 words) in this dataset is much longer than that of other three datasets. The UNC contains 142,209 expressions for 50,000 objects in 19,994 images. The UNC+ is composed of 141,564 expressions for 49,856 objects in 19,992 images and focuses on appearance-based descriptions. Similar to \cite{hu2016segmentation,liu2017recurrent}, the overall Intersection-over-Union (IoU), which calculates total intersection region over total union regions of all test images, is adopted to evaluate our model. We also report the precision at difference threshold, \emph{i.e.}, \emph{P@X} ($X\in{\{0.5,0.6,0.7,0.8,0.9\}}$).

\textbf{Implementation Details}. Similar to \cite{liu2017recurrent,li2018referring}, the commonly used DeepLab ResNet-101 \cite{chen2017deeplab}, which is pre-trained on Pascal VOC \cite{everingham2010pascal}, is utilized as our CNN backbone to extract different level visual features. This backbone is fixed during training and testing. The input images are resized to $320\times{320}$. For language encoding, we first keep the maximum length of each expression as 20. The network is trained using Adam optimizer with an initial learning rate of $2.5e^{-4}$ and a weight decay of $5e^{-4}$. We apply a polynomial decay with power of 0.9 to the learning rate. For the feature dimensions, we set $D_{v}=D_{w}=1000$, $D_{o}=500$.

%For fair comparison with prior works, we report both segmentation results with and without DenseCRF \cite{krahenbuhl2011efficient} post-processing.

\vspace{-0.3cm}
\subsection{Comparison with the State-of-the-arts}
\vspace{-0.3cm}

\begin{table}[!hbt]
	\vspace{-0.22cm}
	\caption{Comparisons with the state-of-the-art methods on four datasets. ``$\star$'' denotes the results are post-processed with DenseCRF.}
	\vspace{-0.25cm}
	\scriptsize
	\label{table:comparison_iou}
	\renewcommand{\arraystretch}{1.25}
	\centering
	\setlength{\tabcolsep}{2.3pt}
	\begin{tabular}{l|ccc|ccc|c|c}
        \hline
		\multirow{2}{*}{\textbf{Methods}}  & \multicolumn{3}{c|}{\textbf{UNC}}  & \multicolumn{3}{c|}{\textbf{UNC+}}  &  \textbf{G-Ref} & \textbf{ReferIt} \\
		
		\cline{2-9}
		&val  &testA  &testB  &val  &testA  &testB  &val  &test \\
		\hline
		\hline
		LSTM-CNN (2016) \cite{hu2016segmentation}           & -          & -         & -         & -         & -          & -         &28.14   & 48.03 \\
		$\rm{RMI^{\star}}$ (2017) \cite{liu2017recurrent}   &45.18       &45.69      &45.57      &29.86      &30.48       &29.50      &34.52   & 58.73 \\
		%DMN (2018) \cite{margffoy2018dynamic}               &49.78       &54.83      &45.13      &38.88      &44.22       &32.29      &36.76   & 52.81 \\
		%KWAN (2018) \cite{shi2018key}                       & -          & -         & -         & -         &-           & -         &36.92   & 59.19 \\
		$\rm{RRN^{\star}}$ (2018) \cite{li2018referring}    &55.33       &57.26      &53.95      &39.75      &42.15       &36.11      &36.45   & 63.63 \\
		ASGN (2019) \cite{qiu2019referring}                 &50.46       &51.20      &49.27      &38.41      &39.79       &35.97      &41.36   & 60.31  \\	
		$\rm{CMSA^{\star}}$ (2019) \cite{ye2019cross}       &58.32       &60.61      &55.09      &43.76      &47.60       &37.89      &39.98   & 63.80 \\
        $\rm{DCLSTM^{\star}}$ (2020) \cite{ye2020dual}      &59.04       &60.74      &56.73      &44.54      &47.92       &39.73      &41.77   & 63.92 \\
        CGAN (2020) \cite{luo2020cascade}                   &59.25       &62.37      &53.94      &46.16      &51.37       &38.24      &46.54   & -     \\
        QRN (2020) \cite{2020Query}                         &59.75       &60.96      &58.77      &48.23      &52.65       &40.89      &42.11   & 65.22 \\
		STEP (2019) \cite{chen2019see}                      &60.04       &63.46      &57.97      &48.19      &52.33       &40.41      &46.40   & 64.13 \\
        $\rm{BRINet^{\star}}$ (2020) \cite{hu2020bi}        &61.35       &63.37      &59.57      &48.57      &52.87       &42.13      &48.04   &63.46 \\
        %CMPC + DCRF \cite{2020cmpc}               &61.36       &64.53      &59.64      &49.56      &53.44       &43.23      &49.05   &65.53 \\
        %LSCM + DCRF \cite{hui2020linguistic}      &61.47       &64.99      &59.55      &49.34      &53.12       &43.50      &48.05   &66.57 \\
		
		\hline
		Ours                      &61.69  &64.33  &59.76  &49.38  &53.47  &42.05  &45.40   &64.67 \\
        $\rm{Ours^{\star}}$       &62.03  &64.70  &60.10  &49.62  &53.85  &42.22  &45.65   &64.72 \\
		\hline
	\end{tabular}
\vspace{-0.45cm}
\end{table}

%Table \ref{table:comparison_iou} shows the comparison results in overall \emph{IoU} between our method and previous state-of-the-art methods. As illustrated in Table \ref{table:comparison_iou}, our method is outperforms previous methods on four benchmark datasets.

We conduct experiments on four benchmark datasets and compare the corresponding results with previous methods in Table \ref{table:comparison_iou}. Note that the symbol ``$\star$'' denotes the segmentation results of the corresponding method are post-processed with DenseCRF \cite{krahenbuhl2011efficient}. It can be observed that the performance of our method outperforms all previous methods on UNC and UNC+. Compared with the best method BRINet, our method achieves $0.68\%$, $1.33\%$ and $0.5\%$ improvement on val, testA and testB sets of UNC, $1.05\%$, $0.98\%$ and $0.09\%$ improvement on val, testA and testB sets of UNC+. In particular, the expression of UNC+ does not include location words, the experimental results show that our method is robust to align the semantics of objects between two modalities. Since G-Ref tends to describe an object with a long expression, and ReferIt contains stuff categories, they are more challenging than UNC and UNC+. Although our method does not establish complex interactions between two modalities like \cite{chen2019see,luo2020cascade,hu2020bi}, the performance of our method also outperforms most state-of-the-art methods on G-Ref and ReferIt. Thus, the integration of multi-scale context information is helpful to align two modalities and improve segmentation accuracy of objects.

%To further exhibit the superiority of our method on object segmentation, we provide some visualization examples of segmentation results on four datasets in supplementary materials.

\vspace{-0.3cm}
\subsection{Ablation Study}
\vspace{-0.15cm}

% scripts
% ablation study of referring image segmentation
% result on unc val
\begin{table}[!hbt]
	\vspace{-0.5cm}
	\caption{Ablation study on the UNC val set with the visual feature $V_{5}$ (post-processed by DenseCRF).}
	\vspace{-0.25cm}
	\scriptsize
	%\small
	\label{table:ablation_study0}
	\renewcommand{\arraystretch}{1.25}
	\centering
	\setlength{\tabcolsep}{3.9pt}
	\begin{tabular}{l|c|c|c|c|c|c}
		\hline
		\textbf{Method} &\textbf{P@0.5} &\textbf{P@0.6} &\textbf{P@0.7} &\textbf{P@0.8} &\textbf{P@0.9} & \textbf{IoU} \\
		\hline
		\hline
        RRN \cite{li2018referring}            &46.82 &36.47 &25.25 &12.25 &1.67  &46.87 \\

        CMSA \cite{ye2019cross}                &51.95 &43.11 &32.74 &19.28 &4.11  &50.12 \\
        %\hline
        DCLSTM \cite{ye2020dual}               &54.62 &44.20 &30.77 &16.02 &2.56  &50.50 \\
        %\hline

        %\hline
        BRINet \cite{hu2020bi}                 &65.53 &57.46 &46.85 &30.42 &7.28  &56.76 \\
        \hline

		Baseline                                   &48.15     &38.56   &28.11    &16.58      &3.81     &48.13 \\
        %\hline
        Baseline + ATTN                            &50.54     &41.20   &30.13    &17.74      &4.36     &49.50 \\
        %\hline
        Baseline + ATTN + ASPP                     &61.48     &53.61   &43.47    &28.49      &7.41     &55.30  \\
        %\hline
        %Cascaded Dilation Conv.     		        &68.13     &61.33   &51.63    &35.83      &10.21    &58.76 \\
        %\hline
        CMF Module                                 &68.60     &62.35   &52.35    &36.40      &10.23    &59.05 \\
        \hline
        			
	\end{tabular}
\vspace{-0.3cm}
\end{table}

% result on unc val
\begin{table}[!hbt]
	\vspace{-0.3cm}
	\caption{The impact of stacking $3\times{3}$ convolutional layers with different dilated rates in CMF moduel (post-processed by DenseCRF).}
	\vspace{-0.25cm}
	\scriptsize
	%\small
	\label{table:ablation_study1}
	\renewcommand{\arraystretch}{1.25}
	\centering
	\setlength{\tabcolsep}{3.7pt}
	\begin{tabular}{l|c|c|c|c|c|c}
		\hline
		\textbf{Methods} &\textbf{P@0.5} &\textbf{P@0.6} &\textbf{P@0.7} &\textbf{P@0.8} &\textbf{P@0.9} & \textbf{IoU} \\
		\hline
        \hline

		Ours (Conv $1\times{1}$)      &50.54 &41.20 &30.13 &17.74 &4.36 &49.50 \\
        %\hline
        Ours (R = 1)                  &53.45 &44.63 &34.00 &21.14 &5.37 &51.09 \\
        %\hline
        Ours (R = 1,3)                &59.61 &51.46 &41.21 &26.93 &7.20 &54.21 \\
        %\hline
        Ours (R = 1,3,5)              &64.02 &56.42 &46.40 &31.67 &8.76 &56.72 \\
        %\hline
        Ours (R = 1,3,5,7)            &67.66 &60.71 &51.29 &35.32 &9.97  &58.48 \\
        %\hline
        Ours (R = 1,3,5,7,9)          &68.60 &62.35 &52.35 &36.40 &10.23 &59.05 \\
        %\hline
        Ours (R = 1,3,5,7,9,11)       &69.33 &62.91 &53.04 &37.50 &10.97 &59.26 \\
        %\hline
        Ours (R = 1,3,5,7,9,11,13)    &70.00 &63.25 &53.38 &37.07 &10.64 &59.41 \\
				
		\hline

	\end{tabular}
\vspace{-0.1cm}
\end{table}

% result on unc val
\begin{table}[!hbt]
	\vspace{-0.2cm}
	\caption{The impact of integrating fused multi-modal features from different levels with Bi-ConvGRU (post-processed by DenseCRF).}
	\vspace{-0.25cm}
	\scriptsize
	%\small
	\label{table:ablation_study2}
	\renewcommand{\arraystretch}{1.25}
	\centering
	\setlength{\tabcolsep}{3.5pt}
	\begin{tabular}{l|c|c|c|c|c|c}
		\hline
		\textbf{Methods} &\textbf{P@0.5} &\textbf{P@0.6} &\textbf{P@0.7} &\textbf{P@0.8} &\textbf{P@0.9} & \textbf{IoU} \\
		\hline
		\hline
		m=\{$\widetilde{V}_{5}$\}                                                                       &68.60     &62.35     &52.35     &36.40    &10.23      &59.05 \\
        %\hline
        Bi-GRU, m=\{$\widetilde{V}_{5}, \widetilde{V}_{4}$\}                                            &72.06     &65.04     &54.74     &37.04    &10.17      &61.04 \\
        %\hline
        Bi-GRU, m=\{$\widetilde{V}_{5}, \widetilde{V}_{4}, \widetilde{V}_{3}$\}                         &71.78     &65.53     &56.25     &40.75    &13.20      &61.50 \\
        %\hline
        Bi-GRU, m=\{$\widetilde{V}_{5}, \widetilde{V}_{4}, \widetilde{V}_{3}, \widetilde{V}_{2}$\}      &72.55     &66.07     &56.90     &41.42    &13.24      &62.03 \\
        %\hline
        %GRU, m=\{$\widetilde{V}_{5}, \widetilde{V}_{4}, \widetilde{V}_{3}, \widetilde{V}_{2}$\}         & & & & & & \\
        %\hline
        %GRU, m=\{$\widetilde{V}_{2}, \widetilde{V}_{3}, \widetilde{V}_{4}, \widetilde{V}_{5}$\}         &72.22     &65.89     &56.44      &40.50    &13.34      &61.72 \\
        %\hline
        %Concatenate, m=\{$\widetilde{V}_{2}, \widetilde{V}_{3}, \widetilde{V}_{4}, \widetilde{V}_{5}$\}    & & & & & & \\
				
		\hline	

	\end{tabular}
\vspace{-0.5cm}
\end{table}

%We perform ablation studies on UNC val set to verify the effectiveness of our proposed Cascaded ASPP Fusion Module, the corresponding experimental results are summarized in Table \ref{table:ablation_study0} and Table \ref{table:ablation_study1}.

In order to verify the effectiveness of our CMF module, we first conduct ablation studies without Bi-ConvGRU on UNC val set. As \cite{hu2016segmentation,liu2017recurrent,li2018referring}, we take $V_{5}$ from \emph{Res5} as the visual feature and fuse it with the linguistic feature. Here we consider three variants of CMF module: (1) Baseline: Following \cite{hu2016segmentation,li2018referring}, this model uses the last hidden state of LSTM as the holistic representation of an expression. We use Hadamard product followed by a convolutional layer with filter size 1 to fuse two modalities. (2) Baseline + Attention (ATTN): This model uses visual feature as a guidance to adaptively learn a robust linguistic representation with an attention mechanism. (3) Baseline + ATTN + ASPP: This model introduces an original ASPP model on the fused multi-modal feature, similar to \cite{luo2020cascade,hu2020bi}. The experimental results are summarized in Lines 5-8 of Table \ref{table:ablation_study0}. It can be observed that CMF module brings $10.92\%$ improvement on baseline model, and brings $3.75\%$ improvement compared with directly using original ASPP. Besides, we also compare CMF module with previous methods only using visual features of $V_{5}$, which are shown in Line 1-4 of Table \ref{table:ablation_study0}. We observe that CMF module outperforms the second best by $2.29\%$ when only use the visual feature of $V_{5}$.

We further study the impact of $3\times{3}$ convolutional layers with gradually increased dilated rate R in the cascaded branch, the results are shown in Table \ref{table:ablation_study1}. We observe that by stacking convolutional layers with bigger dilated rates, the IoU is increased clearly. When we stack five convolutional layers with corresponding dilated rates as $\rm{R}=\{1,3,5,6,9\}$, the IoU is improved by $10.08\%$ compared with the model only uses an $1\times{1}$ convolutional layer. The improvement slows down by stacking more convolutional layers of bigger dilated rates, thus in our model we stack five convolutional layers with dilated rates as $\rm{R}=\{1,3,5,7,9\}$. Finally, we conduct experiments to verify the relative contributions of the introduced Bi-ConvGRU for the fusion of multi-modal feature from different levels. As presented in Table \ref{table:ablation_study2}, when we gradually integrate more fused multi-modal features from lower levels, the segmentation performance is increased.

\vspace{-0.3cm}
\section{Conclusions}
\label{sec:conclusion}
\vspace{-0.3cm}

We have proposed an effective cascaded multi-modal fusion (CMF) module for referring image segmentation. It stacks multiple atrous convolutional layers in parallel and further introduces a cascaded branch to fuse visual and linguistic features using these layers with gradually increased dilated rates. The model can iteratively integrate multi-scale context and facilitate the alignment of two modalities during multi-modal fusion process. We perform experiments on four benchmark datasets and achieve state-of-the-art performance.

\small
\bibliographystyle{unsrt}
\bibliography{strings,refs}

\end{document}